\documentclass[journal]{IEEEtran}
\usepackage[utf8]{inputenc}
\usepackage{CJKutf8}
\usepackage{amsmath}
\usepackage{amssymb}
\usepackage{booktabs}
\usepackage{color}
\usepackage[noadjust]{cite}

\DeclareMathOperator*{\argmax}{arg\,max}

\usepackage{xparse}
\ExplSyntaxOn
\NewDocumentCommand\cb{}
{
    \tl_if_in:cnTF {f@series} {b}
    { \mdseries }
    { \bfseries }
}
\ExplSyntaxOff

\ifCLASSINFOpdf
  \usepackage[pdftex]{graphicx}
  \graphicspath{{figures/}{../pdf/}{../jpeg/}}
  \DeclareGraphicsExtensions{.pdf,.jpeg,.png}
\else
\fi

\begin{document}
\title{Domain-shift Conditioning using Adaptable Filtering via Hierarchical Embeddings for Robust Chinese Spell Check}

\author{Minh~Nguyen~\IEEEmembership{Student Member, IEEE},
        Gia~H.~Ngo,
        and Nancy~F.~Chen~\IEEEmembership{Senior Member,~IEEE}%
\thanks{
© 2021 IEEE.
Personal use of this material is permitted.
Permission from IEEE must be obtained for all other uses, in any current or future media, including reprinting/republishing this material for advertising or promotional purposes, creating new collective works, for resale or redistribution to servers or lists, or reuse of any copyrighted component of this work in other works.\protect\\
Nancy F. Chen is the corresponding author (nancychen@alum.mit.edu).}%
}

\markboth{Journal of \LaTeX\ Class Files,~Vol.~14, No.~8, August~2021}%
{Shell \MakeLowercase{\textit{et al.}}: Bare Demo of IEEEtran.cls for IEEE Journals}

\maketitle

\begin{abstract}
Spell check is a useful application which processes noisy human-generated text.
Spell check for Chinese poses unresolved problems due to the large number of characters, the sparse distribution of errors, and the dearth of resources with sufficient coverage of heterogeneous and shifting error domains.
For Chinese spell check, filtering using confusion sets narrows the search space and makes finding corrections easier.
However, most, if not all, confusion sets used to date are fixed and thus do not include new, shifting error domains.
We propose a scalable adaptable filter that exploits hierarchical character embeddings to (1) obviate the need to handcraft confusion sets, and (2) resolve sparsity problems related to infrequent errors.
Our approach compares favorably with competitive baselines and obtains SOTA results on the 2014 and 2015 Chinese Spelling Check Bake-off datasets.
\end{abstract}

\IEEEpeerreviewmaketitle

\section{Introduction}\label{sec:intro}
\begin{figure}[h]
\centering
\includegraphics[scale=0.4]{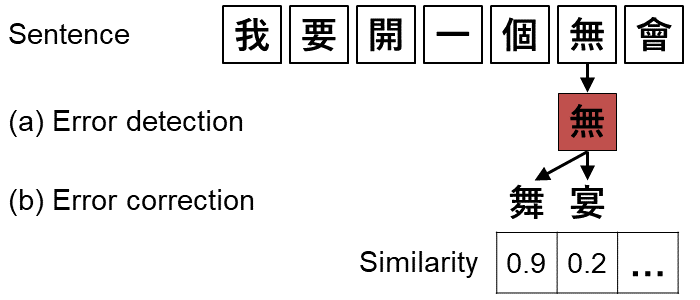}
\caption{\label{fig:intro}\textit{Spell check for Chinese (a) Error detection (b) Error correction (suggesting correction candidates)
}}
\end{figure}

\IEEEPARstart{S}{pell} check is a common task in processing written text, as spell checkers are an integral component in text editors and search engines.
A spell checker must identify erroneous words/characters and suggest candidates for correction (Figure~\ref{fig:intro}).
Despite its utility, spell check for Chinese presents unresolved challenges stemming from the language's numerous characters (up to 10k), the sparsity of errors, and the scarcity of linguistic resources with sufficient coverage of heterogeneous and shifting error domains.

\subsection{Background}\label{ssec:background}
Spell check is more challenging for Chinese than for languages like English as there are more options (characters) to correct an erroneous character.
There are up to 10k common Chinese characters compared to only 26 characters for English.
The sparse and shifting errors also make training a good spell checker challenging.
Errors belong to different domains depending on whether the text with errors is typed or hand-written.
User demographics and degree of formality may also shift the domain of errors.
For example, a spell checker trained using errors collected from essays hand-written by second language learners for academic exams may not generalize well to text typed by native language users on social media platforms.
\begin{figure}[ht]
\centering
\includegraphics[scale=0.6]{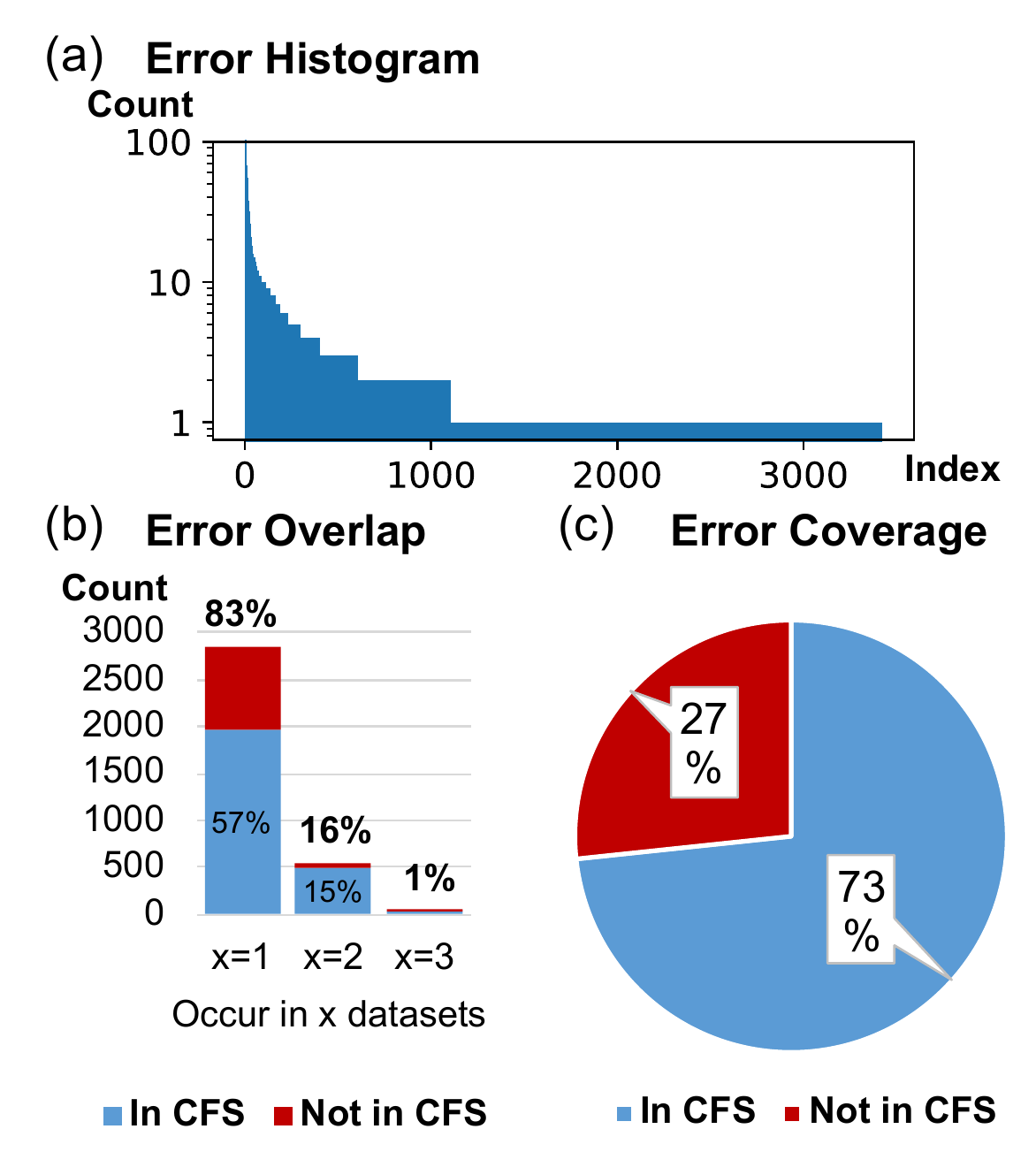}
\caption{\label{fig:sparsity}\textit{Characterizing error domains in datasets from 3 consecutive CSC Bake-offs (a) Error sparsity; More than 2k types of error occur only once across 3 years.
(b) Domain-shift from year to year; 83\% of types of error do not overlap across 3 years.
(c) Domain-shift reduces coverage of given confusion sets (CFS) to only 73\%.
}}
\end{figure}

Figure~\ref{fig:sparsity} illustrates the challenges from three perspectives using datasets from 3 consecutive years of the Chinese spell check (CSC) Bake-offs~\cite{wu2013chinese,yu2014overview,tseng2015introduction}.
Figure~\ref{fig:sparsity}a shows the histogram of error types in the 3 training sets, in which there is a long tail with more than 2000 error types that occur only once (error sparsity).
Figure~\ref{fig:sparsity}b shows the numbers of error types that occur in 1, 2, or all 3 datasets.
The error domains differ vastly between datasets (i.e.\ domain-shift~\cite{ben2007analysis,quionero2009dataset}) with only 1\% of error types common across all 3 datasets.
Most types of error occur exclusively in one dataset (83\%).

Confusion sets~\cite{wu2010reducing} are often used to tackle error sparsity.
The confusion set of a character consists of phonologically or morphologically similar characters.
For example, the confusion set of \begin{CJK*}{UTF8}{bsmi}無\end{CJK*} can be \{\begin{CJK*}{UTF8}{bsmi}吾,嫵,舞\end{CJK*}\}.
Due to their similarity, \begin{CJK*}{UTF8}{bsmi}無\end{CJK*} can be mistaken for any character in its confusion set.
As confusion sets define characters that are likely substitution errors, they can be used to filter out unlikely corrections even for characters unseen in the training data.

However, confusion sets~\cite{wu2013chinese} are often handcrafted and may be susceptible to domain-shift.
As error domains in the data diverge from error domains captured by the handcrafted confusion sets, the number of errors covered by the confusion sets drop to 73\%, leaving a quarter of the errors at risk of mis-detection (Figure~\ref{fig:sparsity}c).
The reduced coverage may also limit the ability to detect infrequent spelling errors, which is in reality a sizable amount of the errors (more than 60\% of errors only occur once), given the extremely long tail in the error distribution (see Figure~\ref{fig:sparsity}).

Constructing confusion sets covering all error domains is nontrivial because there are numerous different domains with various error patterns.
Besides the difference between typed and hand-written text, even within typed text, error patterns differ depending on the input method (IME) used.
Using phonetic-based IMEs like Sogou Pinyin requires inputting pronunciation, so errors often are caused by pronunciation similarity.
In contrast, using morphology-based IMEs like Wubi requires inputting characters' components, so errors are often caused by morphological similarity.
Covering multiple domains is required for accurate spell check because even the same user may commit errors in different domains when using different IMEs (e.g.\ a user who typically uses morphology-based IMEs might switch to phonetic-based IMEs when she forgets how to write but knows how to pronounce a Chinese character).

Keeping the confusion sets up-to-date is as difficult as constructing them as error patterns evolve or new domains emerge.
There is much work attempting to expand or construct confusion sets to increase coverage~\cite{wang2013automatic,chu2014ntou,xie2015chinese}.
However, these methods involve human experts handcrafting the function that measures similarity between characters~\cite{xie2015chinese} or handcrafting how similar characters are gathered~\cite{wang2013automatic,chu2014ntou}.
It is difficult to scale methods with handcrafted rules to capture shifting error domains.

\subsection{Proposed Solution}\label{ssec:proposed_solution}
Due to the sparsity of spelling errors (many errors observed only once in the training data, or possibly not at all) and the large number of Chinese characters, training a spell check model that directly predicts spell error corrections is difficult.
We propose a spell check model with an adaptable filter that refines its prediction by considering phonological and morphological similarity between characters.
The adaptable filter is constructed using a similarity function learned from data using hierarchical character embeddings~\cite{nguyen2019hierarchical}.
Since similarity between characters is one of the primary factors causing substitution errors~\cite{liu2011visually}, the refinement would make spell check more accurate.
Although character embeddings have been used as model input in prior work~\cite{nguyen2019hierarchical,meng2019glyce}, as far as we know, our work is the first that uses character embeddings to filter model output to improve accuracy.
Unlike handcrafted confusion sets, our filter can be trained using errors in the training data so it is less affected by domain-shift.
Experimental results show that the proposed model with adaptive filtering is more accurate than baseline models, obtaining SOTA results on the 2014 and 2015 Chinese Spelling Check Bake-off datasets.

\section{Related Work}\label{sec:related}
Spell check involves error detection and correction to which there are several approaches.
Error detection was tackled using language model~\cite{chang1995new,yeh2013chinese,huang2014chinese}, conditional random field~\cite{wang2014nctu,gu2014introduction}, graph-based algorithm~\cite{jia2013graph,jia2014joint,xin2014improved,zhao2017hybrid}, or sequence tagging model~\cite{xiong2014extended,duan2019chinese,xie2019automatic}.

Most approaches to error correction~\cite{yu2013candidate,li2018chinese} share two common points.
First, they use confusion sets~\cite{hsieh2013introduction,yu2014chinese} to filter out unlikely correction candidates or to generate candidates for beam-search decoding~\cite{bao2020chunk}.
The confusion sets are often constructed once using hand-crafted similarity functions~\cite{zhang2015hanspeller++} and stay fixed thereafter.
Second, most approaches involve data augmentation~\cite{liu2013hybrid,wang2018hybrid} or transfer learning~\cite{hong2019faspell,xie2019automatic,cheng2020spellgcn} to compensate for the limited size of CSC training sets.

Our work is similar to~\cite{xie2019automatic,hong2019faspell,cheng2020spellgcn}, although our filtering is different.
Xie et al.~\cite{xie2019automatic} used fixed filtering while our filtering is adaptable.
Hong et al.~\cite{hong2019faspell} filtering can be fine-tuned albeit manually while our filter can be fine-tuned automatically using training data.
While Cheng et al.~\cite{cheng2020spellgcn}'s filtering exploits only similarity between characters, our approach also exploits character structural similarity by using hierarchical embeddings.
Our filter adaptation is a form of supervised domain adaptation~\cite{daume2007frustratingly} in which labeled errors help capture additional error domains that are missed by the given confusion sets.
Although domain adaptation is usually framed as semi-supervised learning where target domain examples are unlabeled~\cite{glorot2011domain,schnabel2014flors}, semi-supervised adaptation has yet to outperform supervised adaptation in terms of performance improvement~\cite{blitzer2006domain,dredze2007frustratingly,ruder2018strong}.
Thus, through adaptation, our approach may generalize to more error domains and scale more elegantly since it does not require manual tuning.

\section{Approach}\label{sec:method}
\begin{figure}[ht]
\centering
\includegraphics[width=\linewidth]{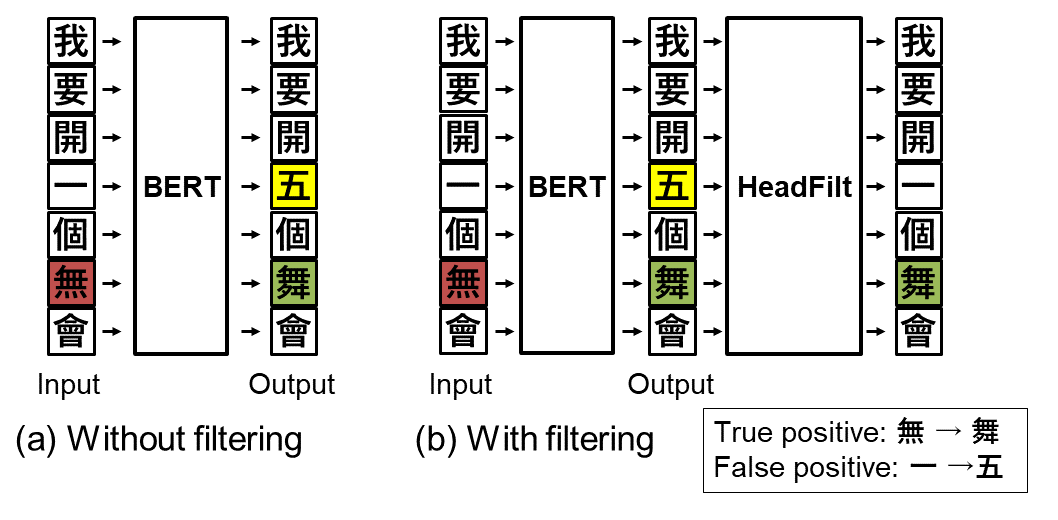}
\caption{\label{fig:training}\textit{Advantage of filtering
(a) Prediction without filtering.
(b) Prediction with filtering. BERT's mistake is corrected by filtering.
Hierarchical Embedding Adaptable Filter (HeadFilt) is described in Section~\ref{ssec:filter_training}.
Red: error, Green: correction, Yellow: false positive.
}}
\end{figure}

The model consists of a pre-trained masked language model (BERT~\cite{devlin2019bert}) and our proposed Hierarchical Embedding Adaptable filter (HeadFilt).
Leveraging pre-trained language model (LM) has led to successes in numerous NLP tasks~\cite{ramachandran2017unsupervised,peters2018deep,radford2018improving,howard2018universal}.
However, predicting corrections using only the masked LM (Figure~\ref{fig:training}a) is challenging due to inadequate training data and the large number of Chinese characters.
To boost accuracy, LM's predictions are filtered using HeadFilt (Figure~\ref{fig:training}b).

Let the input sequence be $\mathbf{X_*}$, where $X_i$ denotes the $i$-th character in the sequence. All unique characters across all sequences belong to the vocabulary $\{X\}$.
Given $\mathbf{X_*}$, the masked LM outputs $\mathbf{\tilde{Y}_*}$, whereby $\mathbf{\tilde{Y}}_i=\{\tilde{Y}_{ik}\}$ is the distribution over corrections at $i$.
The predicted correct character at $i$ before filtering is $\tilde{Z}_i$ (Equation~\ref{eq:nofilt}).

For each input character $X_i$, HeadFilt computes a similarity vector of size $N$, $\mathbf{S}_i$, which is the similarity between $X_i$ and the set of all $N$ characters in the vocabulary (i.e.\ $\{X\}$).
HeadFilt is described in Section~\ref{ssec:filter_intro}.
The similarity vector $\mathbf{S}_i$ is element-wise multiplied with $\mathbf{\tilde{Y}}_i$ to filter out unlikely corrections.
The predicted correct character at position $i$ after filtering is $\tilde{Z}^{\text{filt}}_i$ (Equation~\ref{eq:filt}).
\begin{align}
    \mathbf{\tilde{Y}}_* &= \text{MaskLM} (\mathbf{X_*}) \\
    \tilde{Z}_i &= \argmax_{k} \tilde{Y}_{ik}\label{eq:nofilt} \\
    \mathbf{S}_i &= \text{HeadFilt} (X_i; \{X\}) \\
    \tilde{Z}^{\text{filt}}_i &= \argmax_{k} (\mathbf{\tilde{Y}}_{i} \odot \mathbf{S}_i)_{k} \label{eq:filt}
\end{align}

\subsection{Adaptable Filter with Hierarchical Embeddings}\label{ssec:filter_intro}
\begin{figure}[h]
\centering
\includegraphics[width=\linewidth]{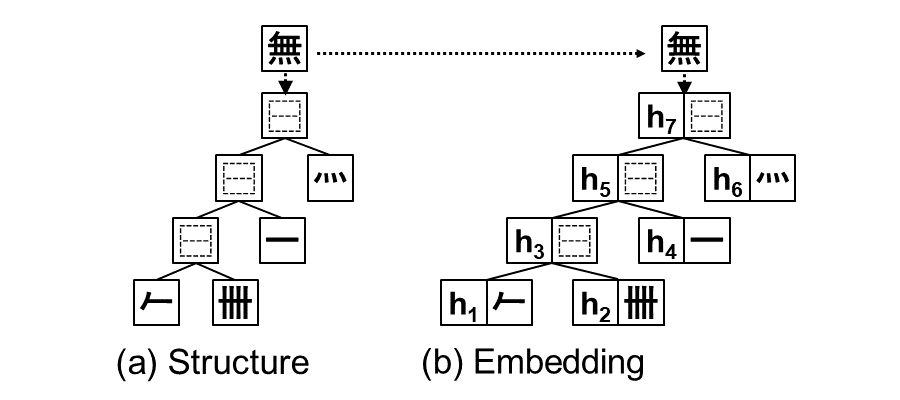}
\caption{\label{fig:emb}\textit{Hierarchical embedding
(a) Character as a tree with sub-character components as leaves. Internal nodes denote the relative spatial positions of the character components.
(b) Hierarchical embedding ($\mathbf{h_7}$) constructed from tree
}}
\end{figure}

Filtering out unlikely candidates using confusion sets can significantly boost accuracy~\cite{hsieh2013introduction,yu2014chinese,zhang2015hanspeller++}.
However, under domain-shift, fixed confusion sets could miss errors from new domains.
We propose an adaptable filtering model named HeadFilt which can be fine-tuned using errors from new domains to alleviate the domain-shift problem.
Filter fine-tuning is a form of supervised domain adaptation as it minimizes the domain-shift between errors captured by the given confusion sets and the errors in the training data.
HeadFilt estimates likelihood of correction candidates using distances in the embedding space between characters. 
HeadFilt uses hierarchical character embeddings~\cite{nguyen2019hierarchical} which have inductive bias of the character structure so as to extend filtering to phonologically and morphologically characters unseen in the training data, thus addressing the error sparsity problem.
Hierarchical character embedding is obtained by applying TreeLSTM~\cite{tai2015improved,zhu2015long} on the tree structure of the character.
Figure~\ref{fig:emb} shows the tree structure of a character and its hierarchical embedding ($\mathbf{h_7}$).

\begin{figure}[ht]
\centering
\includegraphics[width=\linewidth]{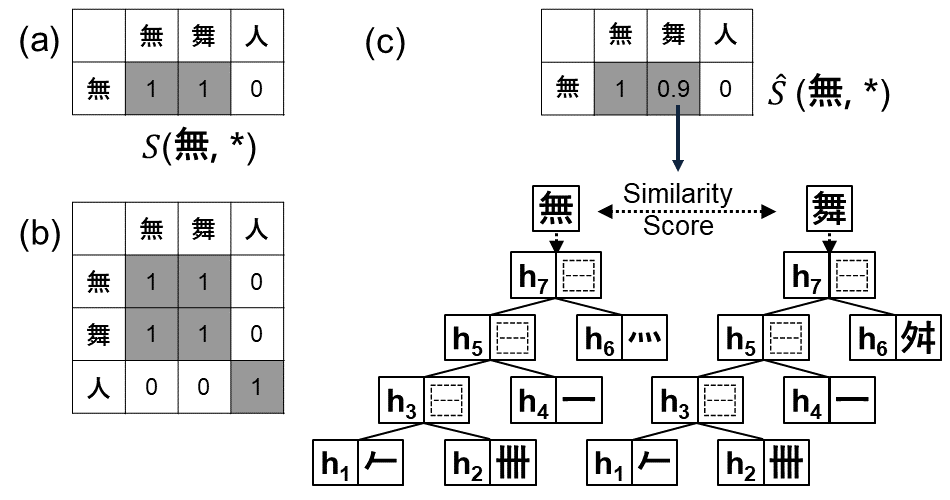}
\caption{\label{fig:confusion}\textit{(a) Similarity vector constructed from the confusion set of character \begin{CJK*}{UTF8}{bsmi}無\end{CJK*}
(b) All confusion sets
(c) Similarity vector constructed by HeadFilt
}}
\end{figure}

A (fixed) confusion set of a character can be represented as a binary similarity vector ($\mathbf{S}$ in Figure~\ref{fig:confusion}a) with a $1$ for each character in the set that is deemed possibly confused with the given character.
In contrast, the similarity vector ($\mathbf{\widehat{S}}$ in Figure~\ref{fig:confusion}c) produced by HeadFilt has real-valued scores estimated using the structure of the characters.
Combining all the confusion sets (i.e.\ concatenating all the binary vectors) results in a similarity matrix (Figure~\ref{fig:confusion}b).

Let the hierarchical embeddings of character $a$ and $b$ be $\mathbf{h_a}$ and $\mathbf{h_b}$.
The HeadFilt similarity between $a$ and $b$, $\widehat{S}(a, b)$, is estimated using Equation~\ref{eq:sim}.
The constants $\beta$ and $m$ in Equation~\ref{eq:sim} are the scaling factor and the margin respectively.
Intuitively, $\widehat{S}(a, b)$ in Equation~\ref{eq:sim} is close to 1 if the L2 distance ($d_{ab}$) between the two embedding vectors is smaller than the margin $m$ and is close to 0 otherwise.
Let $\mathbf{\widehat{S}}_i$ be the HeadFilt similarity vector of the character $X_i$.
Vector $\mathbf{\widehat{S}}_i$ is the concatenation of similarity scores of $X_i$ and all other $N$ characters ($c_1, c_2, \dots c_N$) in the vocabulary. (Equation~\ref{eq:sim_vec}).
\begin{align}
d_{ab} &= \Big\lVert \mathbf{h_a} / \lVert \mathbf{h_a} \lVert - \mathbf{h_b} / \lVert \mathbf{h_b} \lVert \Big\lVert \\
\widehat{S}(a, b) &= \frac{1}{1 + \exp(\beta \times (d_{ab} - m))}\label{eq:sim} \\
\mathbf{\widehat{S}}_i &= [\widehat{S}(X_i, c_1), \dots \widehat{S}(X_i, c_N)]\label{eq:sim_vec}
\end{align}

In addition, let $\mathbf{S}_i$ be the confusion set similarity vector of $X_i$. The prediction after filtering ($\tilde{Z}^{\text{filt}}_i$) is:
\begin{align}
    \mathbf{\tilde{Y}}^{\text{filt}}_i &= \mathbf{\tilde{Y}}_i \odot \mathbf{S}_i \;\;\text{(using fixed confusion set)} \\
    \mathbf{\tilde{Y}}^{\text{filt}}_i &= \mathbf{\tilde{Y}}_i \odot \mathbf{\widehat{S}}_i \;\;\text{(using HeadFilt)}
\end{align}

\subsection{Adaptable Filter Training}\label{ssec:filter_training}
HeadFilt is trained by minimizing the contrastive loss~\cite{hadsell2006dimensionality} in Equation~\ref{eq:contrast} below.
Minimizing the loss forces L2 distances between similar characters to be within the margin $m$ and L2 distances between dissimilar characters to be greater than the margin $m$.
Let $S(a, b)$ be the observed similarity of two characters $a$ and $b$.
\begin{align}
L_c &= \sum_{a, b} [ S(a, b) \times \max (0, d_{ab} - m) \nonumber \\
    &+ (1 - S(a, b)) \times \max (0, m - d_{ab}) ] \label{eq:contrast}
\end{align}

There are two sources of examples for training the filter: i.e.\ the positive and negative examples from the confusion sets and the positive examples observed in the training data.
A positive example is an error pair $(a, b)$ where $a$ is mistaken for $b$ or vice-versa.
Thus, all pair $(a, b)$ in the same confusion set and all errors $(a, b)$ in the training data are positive examples.
Other character pairs are considered as negative examples.
For positive examples, $a$ and $b$ are similar and $S(a, b) = 1$, while for negative examples, $a$ and $b$ are dissimilar and $S(a, b) = 0$.

The objective of the optimization in Equation~\ref{eq:contrast} is to ensure that (1) the distances between similar characters are smaller than the margin $m$ and (2) the distances between dissimilar characters are larger than the margin $m$.
When a and b are similar, $S(a, b) = 1$, the loss contributed by $a$ and $b$ is $max(0, d_{ab} – m)$.
Minimizing the loss will push the embeddings of $a$ and $b$ closer to one another if only $d_{ab} > m$.
When a and b are dissimilar, $S(a, b) = 0$, the loss contributed by $a$ and $b$ is $max(0, m - d_{ab})$.
Minimizing the loss will push the embeddings of $a$ and $b$ further apart if only $d_{ab} < m$.

HeadFilt is trained in two steps:
(1) imitating the given confusion sets, where characters in the same confusion sets are cast as positive examples and characters not in the same confusion sets are cast as negative examples.
(2) domain-shift conditioning (adaptation), where error pairs observed in the training data are further added as positive examples and the filter is trained further using this larger set of examples.
Combining examples from both source and target domains for supervised adaptation was also done in~\cite{daume2007frustratingly}.
The effect of domain adaptation is shown in Section~\ref{ssec:ablation}.

To calculate $\widehat{S}(a, b)$ (Equation~\ref{eq:sim}), we need the values of $m$ and $\beta$.
In our experiments, we set $m = 0.4$.
Besides, we need to set $\beta$ so that when $a$ and $b$ are dissimilar, the probability of predicting $b$ as the correction for $a$ or vice-versa is very small (less than chance) as shown in Equation~\ref{eq:beta_true}.
\begin{align}
    &\frac{1}{1 + \exp(\beta \times (d_{ab} - m))} \leq \frac{1}{N} , \nonumber \\
    &\forall a, b \;\;\text{s.t.}\;\; S(a, b) = 0 \label{eq:beta_true} \\
    &\beta \geq \frac{\ln{(N - 1)}}{d_{*} - m} \label{eq:beta_approx}\\
    &d_{*} = \frac{\sum_{a,b} (1 - S(a, b)) * d_{ab}}{\sum_{a,b} (1 - S(a, b))}
\end{align}
However, solving Equation~\ref{eq:beta_true} for all pairs $(a, b)$ (including unobserved pairs) is intractable.
Thus, we approximate Equation~\ref{eq:beta_true} by Equation~\ref{eq:beta_approx} using positive and negative examples in the training data.
Setting $\beta$ according to Equation~\ref{eq:beta_approx} ensures that on average when $a$ and $b$ are dissimilar, it is very unlikely to predict $b$ as the correction for $a$.

\section{Experiments}\label{sec:exp}
\subsection{Data}\label{ssec:data}
We used datasets from the 2013~\cite{wu2013chinese}, 2014~\cite{yu2014overview}, and 2015~\cite{tseng2015introduction} Chinese Spell Check Bake-offs.
For each dataset, we only used its training set to fine-tune the models (BERT and HeadFilt).
Table~\ref{tbl:data_stat} shows the statistics of the datasets.
Unlike 2014 and 2015, there are two test sets in 2013: one for error detection and one for error correction.
\begin{table}[ht]
\normalsize
\centering
\begin{tabular}{llrr}
Dataset    & With Error & Length \\
\midrule
2013 Train & 350 / 700 & 41.8 \\
2013 Test (Detection) & 300 / 1000 & 68.7 \\
2013 Test (Correction) & 996 / 1000 & 74.3 \\
\midrule
2014 Train & 3432 / 3435 & 49.6 \\
2014 Test  & 529 / 1062 & 50.0 \\
\midrule
2015 Train & 2339 / 2339 & 31.3 \\
2015 Test  & 550 / 1100 & 30.6 \\
\end{tabular}
\caption{\label{tbl:data_stat}\textit{Amount of data (sentences) in the datasets.}}
\end{table}

We used the confusion sets~\cite{liu2011visually} provided by the 2013 Bake-off for filtering in our baseline model and for training the HeadFilt model.
As the confusion sets were created in 2011, it may not cover newer error domains (i.e.\ domain-shift).
\begin{table}[ht]
\normalsize
\centering
\begin{tabular}{lrr}
Training set & Errors covered by confusion sets \\
\midrule
2013 & 252/269  (93.68\%) \\
2014 & 1641/2197 (74.69\%) \\
2015 & 1177/1568 (75.06\%) \\
\end{tabular}
\caption{\label{tbl:cfs_coverage}\textit{Number of unique errors in the training sets that are covered by the given confusion sets.}}
\end{table}
Table~\ref{tbl:cfs_coverage} shows the number of errors in the training data that are covered by the confusion sets.
Although coverage is above 90\% for 2013, it drops by 20\% absolute for 2014 and 2015.
This drop could conceivably be due to different data collection methods.
While the 2013 dataset was from written essays by first language learners of Chinese, the 2014 and 2015 datasets were from typed essays by second language learners of Chinese.
Despite the lack of coverage, HeadFilt can adapt to errors in the training sets to filter better.

\subsection{Baselines}\label{ssec:baselines}
The proposed approach has two stages in which the masked LM's output from the first stage is filtered by HeadFilt in the second stage.
We compared the proposed approach against 3 baselines:
\begin{itemize}
    \item[1.] BERT which does not filter predictions.
    \item[2.] FixedFilt which uses the provided fixed confusion sets for filtering.
    \item[3.] GlyceFilt which uses Glyce embeddings~\cite{meng2019glyce} instead of hierarchical embeddings.
\end{itemize}
Although there are many other embedding models for Chinese characters~\cite{shi2015radical,yin2016multi,liu2017learning,yu2017joint,su2017learning,cao2018cw2vec,zhuang2018chinese}, we chose the Glyce embedding model as it is the current SOTA for many different Chinese NLP tasks and is therefore a strong baseline.
Results of winning teams of the Bake-offs are also included:
~\cite{chiu2013chinese} (SMT),
~\cite{yeh2013chinese} (n-gram LM),
National Kaohsiung University of Applied Sciences (KUAS),
~\cite{zhang2015hanspeller++} (HanSpeller),
as well as recent results by~\cite{xie2019automatic} (DPL-Corr),~\cite{hong2019faspell} (FASpell), and~\cite{bao2020chunk} (MERT).

Note that Wang et al.~\cite{wang2019confusionset} and Cheng et al.~\cite{cheng2020spellgcn} used different training and test sets, different evaluation tools, and Simplified Chinese characters instead of the original Traditional Chinese.
Therefore their results are not directly comparable to previously mentioned studies.
We include our best attempt to compare with \cite{wang2019confusionset,cheng2020spellgcn} in Section~\ref{ssec:simp_comparison}.

\begin{table}[t]
\small
\centering
\begin{tabular}{l@{\hspace{1em}}l@{\hspace{.8em}}l@{\hspace{.8em}}l@{\hspace{.8em}}l}
Prediction & Acc. & Pre. & Rec. & F1 \\
\midrule
\multicolumn{5}{c}{2013 Bake-off} \\
\midrule
SMT        & \cb 0.861 & 0.845 & 0.656 & \cb 0.739 \\
n-gram LM  & 0.825 & 0.745 & 0.633 & 0.684 \\
FASpell    & 0.631 & 0.762 & 0.632 & 0.691 \\
\midrule
BERT       & 0.905 & 0.835 & 0.851 & 0.842 \\
FixedFilt  & 0.910 & 0.884 & 0.804 & 0.842 \\
GlyceFilt   & \cb 0.917 & 0.910 & 0.804 & \cb 0.854 \\
HeadFilt   & 0.914 & 0.891 & 0.813 & 0.850 \\
\midrule
\multicolumn{5}{c}{2014 Bake-off} \\
\midrule
KUAS       & \cb 0.719 & 0.914 & 0.484 & 0.633 \\
FASpell    & 0.700 & 0.610 & 0.535 & 0.570 \\
SpellGCN$\dagger$   & --- & 0.583 & 0.545 & 0.563 \\
MERT   & 0.700 & 0.787 & 0.548 & \cb 0.646 \\
\midrule
BERT       & 0.708 & 0.780 & 0.580 & 0.665 \\
FixedFilt  & 0.709 & 0.831 & 0.526 & 0.644 \\
GlyceFilt   & 0.702 & 0.824 & 0.515 & 0.633 \\
HeadFilt   & \cb 0.719 & 0.823 & 0.559 & \cb 0.666 \\
\midrule
\multicolumn{5}{c}{2015 Bake-off} \\
\midrule
HanSpeller & 0.700 & 0.802 & 0.532 & 0.640 \\
DPL-Corr   & 0.709 & 0.767 & 0.600 & 0.673 \\
FASpell    & 0.742 & 0.676 & 0.600 & 0.635 \\
SpellGCN$\dagger$   & --- & 0.710 & 0.640 & 0.673 \\
MERT   & \cb 0.768 & 0.881 & 0.620 & \cb 0.728 \\
\midrule
BERT       & 0.763 & 0.849 & 0.641 & 0.730 \\
FixedFilt  & 0.757 & 0.900 & 0.580 & 0.705 \\
GlyceFilt   & 0.764 & 0.902 & 0.592 & 0.715 \\
HeadFilt   & \cb 0.773 & 0.894 & 0.619 & \cb 0.731 \\
\end{tabular}
\caption{\label{tbl:result_pred}\textit{Traditional Chinese error prediction.
The means of 5 runs with different seeds are reported.
See Section~\ref{ssec:baselines} for list of baselines.
Acc: Accuracy, Pre: Precision, Rec: Recall
}}
\end{table}
\begin{table}[t]
\small
\centering
\begin{tabular}{l@{\hspace{1em}}l@{\hspace{.8em}}l@{\hspace{.8em}}l@{\hspace{.8em}}l}
Correction & Acc. & Pre. & Rec. & F1 \\
\midrule
\multicolumn{5}{c}{2013 Bake-off} \\
\midrule
SMT        & 0.443 & 0.699 & --- & --- \\
n-gram LM  & \cb 0.625 & 0.703  & ---   & ---   \\
FASpell    & 0.605 & 0.731 & --- & --- \\
\midrule
BERT       & 0.489 & 0.597 & --- & --- \\
FixedFilt  & 0.575 & 0.737 & --- & --- \\
GlyceFilt   & 0.537 & 0.712 & --- & --- \\
HeadFilt   & \cb 0.591 & 0.754 & --- & --- \\
\midrule
\multicolumn{5}{c}{2014 Bake-off} \\
\midrule
KUAS       & \cb 0.708 & 0.910 & 0.461 &  0.612 \\
FASpell    & 0.693 & 0.594 & 0.520 & 0.554 \\
SpellGCN$\dagger$   & --- & 0.510 & 0.477 & 0.493 \\
MERT   & 0.681 & 0.774 & 0.510 & \cb 0.615 \\
\midrule
BERT       & 0.650 & 0.740 & 0.464 & 0.570 \\
FixedFilt  & 0.684 & 0.816 & 0.475 & 0.601 \\
GlyceFilt   & 0.681 & 0.811 & 0.472 & 0.597 \\
HeadFilt   & \cb 0.698 & 0.811 & 0.517 & \cb 0.631 \\
\midrule
\multicolumn{5}{c}{2015 Bake-off} \\
\midrule
HanSpeller & 0.691 & 0.797 & 0.514 & 0.625 \\
DPL-Corr   & 0.695 & 0.759 & 0.572 & 0.652 \\
FASpell    & 0.737 & 0.666 & 0.591 & 0.626 \\
SpellGCN$\dagger$   & --- & 0.601 & 0.542 & 0.570 \\
MERT   & \cb 0.746 & 0.873 & 0.576 & \cb 0.694 \\
\midrule
BERT       & 0.689 & 0.812 & 0.492 & 0.613 \\
FixedFilt  & 0.729 & 0.890 & 0.523 & 0.658 \\
GlyceFilt   & 0.730 & 0.891 & 0.525 & 0.661 \\
HeadFilt   & \cb 0.746 & 0.885 & 0.565 & \cb 0.690 \\
\end{tabular}
\caption{\label{tbl:result_corr}\textit{Traditional Chinese error correction.
The means of 5 runs with different seeds are reported.
$\dagger$ Results reported by Bao et al.~\cite{bao2020chunk}.
}}
\end{table}

\subsection{Experimental Setup}
We used tools provided by the CSC Bake-offs~\cite{tseng2015introduction} to evaluate accuracy, precision, recall, and F1 score.
Metrics are calculated at the sentence level, i.e.\ a prediction is correct if all errors in the sentence are predicted correctly.

The models are implemented using PyTorch~\cite{paszke2019pytorch}.
BERT model is trained using cross-entropy loss for 20 epochs using a learning rate of $3e^{-5}$ and a batch size of 16 with AdamW~\cite{loshchilov2018decoupled}.
We repeated this training 5 times using 5 different random seeds and reported the average.
HeadFilt is trained using the contrastive loss~\cite{hadsell2006dimensionality} with Adam~\cite{kingma2015adam}.
We used a learning rate of $3e^{-3}$ and a batch size of 500.
The hierarchical embeddings' dimension is 512.
HeadFilt is trained in two steps (see Section~\ref{ssec:filter_training}).
HeadFilt is first trained to imitate the given confusion sets for 150k steps.
HeadFilt is then further trained using the additional error pairs observed in the training data for 50k steps.

\subsection{Results for Traditional Chinese Data}\label{ssec:results}
For error detection (Table~\ref{tbl:result_pred}), approaches with filtering (FixedFilt, GlyceFilt, and HeadFilt) achieved higher precision and lower recall when compared to the non-filtering baseline (BERT).
HeadFilt was consistently better than the BERT baseline in terms of the accuracy and F1 score, validating the benefit of filtering.
HeadFilt was also consistently better than FixedFilt, showing that adaptation is beneficial.
HeadFilt was as good as if not better than GlyceFilt, suggesting that the inductive bias of hierarchical embeddings is suitable for filtering.
The proposed HeadFilt model also achieves SOTA results for error prediction F1 score for 2014 and 2015. 

For error correction (Table~\ref{tbl:result_corr}), HeadFilt is better than FixedFilt and BERT in all datasets in both accuracy and F1 score.
Whereas filtering lowers recall in error prediction, it leads to higher recall in error correction since filtering lowers the number of correction candidates to be considered.
HeadFilt achieved consistently higher recall than the FixedFilt baseline.
HeadFilt is also better than the SOTA result for F1 score in 2014 and 2015.

\subsection{Results for Simplified Chinese Data}\label{ssec:simp_comparison}

\begin{table}[t]
\normalsize
\centering
\begin{tabular}{lllll}
            & Acc. & Pre. & Rec. & F1 \\
\midrule
\multicolumn{5}{c}{2013 Bake-off} \\
\midrule
SpellGCN    & --- & 0.801 & 0.744 & 0.772 \\
\midrule
BERT        & 0.742 & 1.000 & 0.742 & 0.852 \\
FixedFilt   & 0.732 & 1.000 & 0.732 & 0.845 \\
GlyceFilt   & \cb 0.749 & 1.000 & 0.749 & 0.856 \\
HeadFilt    & \cb 0.749 & 1.000 & 0.749 & \cb 0.857 \\
\midrule
\multicolumn{5}{c}{2014 Bake-off} \\
\midrule
SpellGCN    & --- & 0.651 & 0.695 & 0.672 \\
\midrule
BERT        & 0.733 & 0.800 & 0.621 & 0.699 \\
FixedFilt   & 0.712 & 0.815 & 0.550 & 0.657 \\
GlyceFilt   & 0.707 & 0.833 & 0.518 & 0.639 \\
HeadFilt    & \cb 0.742 & 0.825 & 0.616 & \cb 0.705 \\
\midrule
\multicolumn{5}{c}{2015 Bake-off} \\
\midrule
SpellGCN    & --- & 0.748 & 0.807 & 0.777 \\
\midrule
BERT        & 0.785 & 0.827 & 0.721 & 0.770 \\
FixedFilt   & 0.766 & 0.844 & 0.653 & 0.736 \\
GlyceFilt   & 0.781 & 0.871 & 0.661 & 0.752 \\
HeadFilt    & \cb 0.793 & 0.845 & 0.718 & \cb 0.776 \\
\end{tabular}
\caption{\label{tbl:app_pred_all}\textit{Simplified Chinese error prediction.
The means of 5 runs with different seeds are reported.
}}
\end{table}
Due to differences in the experimental setup, results from Wang et al.~\cite{wang2019confusionset} and Cheng et al.~\cite{cheng2020spellgcn} are not directly comparable to those from prior studies~\cite{xie2019automatic,hong2019faspell}.
First, datasets were converted from Traditional to Simplified Chinese in~\cite{wang2019confusionset,cheng2020spellgcn}, potentially altering the results.
This is because distinct traditional characters could be merged and replaced with the same Simplified character, so a character may be correct in Simplified Chinese but erroneous in Traditional Chinese~\cite{hong2019faspell}.
For example, the character \begin{CJK*}{UTF8}{bsmi}周\end{CJK*} in \begin{CJK*}{UTF8}{bsmi}周末\end{CJK*} (``weekend'') is correct in Simplified Chinese, but it is an error in Traditional Chinese (the correct character is \begin{CJK}{UTF8}{bsmi}週\end{CJK}).
Second, Wang et al.~\cite{wang2019confusionset} evaluated only on sentences with errors from the original test sets using character level metrics instead of evaluating all the test set sentences using sentence level metrics as it was done at the Bake-offs.
Since we are following the standard set up from the Bake-offs, it is tricky to compare fairly with~\cite{wang2019confusionset}.

\begin{table}[t]
\normalsize
\centering
\begin{tabular}{lllll}
           & Acc. & Pre. & Rec. & F1 \\
\midrule
\multicolumn{5}{c}{2013 Bake-off} \\
\midrule
SpellGCN    & ---   & 0.783 & 0.727 & 0.754 \\
\midrule
BERT        & 0.722 & 1.000 & 0.722 & 0.839 \\
FixedFilt   & 0.709 & 1.000 & 0.709 & 0.829 \\
GlyceFilt   & 0.739 & 1.000 & 0.739 & 0.850 \\
HeadFilt    & \cb 0.741 & 1.000 & 0.741 & \cb 0.851 \\
\midrule
\multicolumn{5}{c}{2014 Bake-off} \\
\midrule
SpellGCN    &   --- & 0.631 & 0.672 & 0.653 \\
\midrule
BERT        & 0.721 & 0.794 & 0.597 & 0.681 \\
FixedFilt   & 0.699 & 0.807 & 0.524 & 0.635 \\
GlyceFilt   & 0.700 & 0.830 & 0.504 & 0.627 \\
HeadFilt    & \cb 0.735 & 0.821 & 0.602 & \cb 0.694 \\
\midrule
\multicolumn{5}{c}{2015 Bake-off} \\
\midrule
SpellGCN    &   --- & 0.721 & 0.777 & 0.759 \\
\midrule
BERT        & 0.770 & 0.821 & 0.691 & 0.751 \\
FixedFilt   & 0.745 & 0.835 & 0.610 & 0.705 \\
GlyceFilt   & 0.773 & 0.868 & 0.644 & 0.740 \\
HeadFilt    & \cb 0.785 & 0.842 & 0.702 & \cb 0.765 \\
\end{tabular}
\caption{\label{tbl:app_corr_all}\textit{Simplified Chinese error correction.
The means of 5 runs with different seeds are reported.
}}
\end{table}

However, it is possible to compare with Cheng et al.~\cite{cheng2020spellgcn}'s result since they also evaluated using sentence level metrics.
We attempted to compare against Cheng et al.~\cite{cheng2020spellgcn}'s method (SpellGCN) by training our models with their training data and evaluated on their test data (in Simplified Chinese).
The results in Table~\ref{tbl:app_pred_all} and~\ref{tbl:app_corr_all} show that HeadFilt outperforms SpellGCN in all three years.
HeadFilt was also generally better than GlyceFilt.
Interestingly, the FixedFilt baseline is worse than the baseline without filtering (i.e. BERT).
The confusion sets used in the FixedFilt baseline are obtained by converting the original confusion sets into Simplified script.
The drop in performance of FixedFilt could be because the confusion sets in simplified scripts do not capture the errors well.
In this setup using simplified characters, only 60\% of the errors in the training set are covered by the confusion sets as compared to more than 75\% in the setup using traditional characters (Table~\ref{tbl:cfs_coverage}).
Such a large shift in error domains severely affects model performance of FixedFilt.
In contrast, with domain adaptation, the HeadFilt model still performs well.
Comparing Table~\ref{tbl:result_pred} and~\ref{tbl:result_corr} with Table~\ref{tbl:app_pred_all} and~\ref{tbl:app_corr_all} shows how model performance varies across different experimental setups.

\subsection{Ablation Study}\label{ssec:ablation}
We perform ablation to gauge the contribution of components to the performance of HeadFilt.
We compared the full HeadFilt model against the HeadFilt model without domain adaptation in Section~\ref{sec:adapt}.
This is denoted as ``-Ad'' in Table~\ref{tbl:ablation_pred} and~\ref{tbl:ablation_corr}.
We further assessed the contribution from the hierarchical embeddings in Section~\ref{sec:hier} by replacing hierarchical embeddings with standard embeddings.
This is denoted as ``-Ad -HE''.

\subsubsection{Adaptation Analysis}\label{sec:adapt}
Domain adaptation contributed little to HeadFilt performance in 2013 since removing this step resulted in the same accuracy and F1 score for both error prediction and correction.
This is not surprising considering that 93.68\% of the errors in the training data is already covered by the confusion sets (shown in Table~\ref{tbl:cfs_coverage}), therefore, adaptation only add 17 new pairs of positive examples.
For 2014 and 2015, domain adaptation shows more tangible contribution, especially for error correction.
This result aligns with statistics in Table~\ref{tbl:cfs_coverage}, which shows larger domain-shifts in 2014 and 2015.

\subsubsection{Hierarchical Embedding Analysis}\label{sec:hier}
Besides domain adaptation, the use of hierarchical embeddings also contributed to the performance gain since replacing hierarchical with standard embeddings leads to worse performance for both error prediction and correction.
This trend is observed in all 3 datasets.
Besides, HeadFilt without domain adaptation (``-Ad'') consistently beats FixedFilt, demonstrating the benefit of hierarchical embeddings.
Hierarchical embeddings impose additional constraints (inductive bias) on the embedding space, such that characters with similar morphology are more likely to have similar character embeddings.
This inductive bias could make it more effective to automatically learn which sets of characters are more likely to be mistaken from one another (See Section~\ref{ssec:inductive_bias}).

\begin{table}[t]
\small
\centering
\begin{tabular}{l@{\hspace{1em}}l@{\hspace{.8em}}l@{\hspace{.8em}}l@{\hspace{.8em}}l}
Prediction & Acc. & Pre. & Rec. & F1 \\
\midrule
\multicolumn{5}{c}{2013 Bake-off} \\
\midrule
HeadFilt   & 0.914 & 0.891 & 0.813 &  0.850 \\
-Ad        & \cb 0.915 & 0.894 & 0.813 & \cb 0.851 \\
-Ad -HE    & 0.911 & 0.906 & 0.787 & 0.842 \\
FixedFilt  & 0.910 & 0.884 & 0.804 & 0.842 \\
\midrule
\multicolumn{5}{c}{2014 Bake-off} \\
\midrule
HeadFilt   & \cb 0.719 & 0.823 & 0.559 & \cb 0.666 \\
-Ad        & 0.717 & 0.827 & 0.548 & 0.659 \\
-Ad -HE    & 0.714 & 0.826 & 0.542 & 0.654 \\
FixedFilt  & 0.709 & 0.831 & 0.526 & 0.644 \\
\midrule
\multicolumn{5}{c}{2015 Bake-off} \\
\midrule
HeadFilt   & \cb 0.773 & 0.894 & 0.619 & \cb 0.731 \\
-Ad        & \cb 0.773 & 0.903 & 0.611 & 0.729 \\
-Ad -HE    & 0.765 & 0.906 & 0.592 & 0.716 \\
FixedFilt  & 0.757 & 0.900 & 0.580 & 0.705 \\
\end{tabular}
\caption{\label{tbl:ablation_pred}\textit{Traditional Chinese error prediction ablation.
The means of 5 runs with different seeds are reported.
Ad: Adaptation, HE: hierarchical embedding,
}}
\end{table}

\section{Discussion}
\subsection{Hierarchical Embeddings Enables Automated Domain Adaptation}\label{ssec:inductive_bias}
Section~\ref{ssec:simp_comparison} shows that using confusion sets without sufficient coverage could hurt performance.
Ideally, there should be a confusion set of at least 100 candidates for every character~\cite{liu2011visually}.
The confusion sets used in this work~\cite{liu2009phonological} cover only 5,401 high frequency Traditional Chinese characters and only 30\% of the confusion sets have more than 100 candidates.
Filtering using these fixed confusion sets could miss many spelling errors, especially for text from other domains (e.g.\ Simplified Chinese text or technical documents).
Changes in word usage might also make confusion sets obsolete since past spelling errors could become widely accepted and used and are no longer considered as errors.
For example, in English, ``ingot'' was a spelling error of the word ``lingot''.
However, as the usage of ``ingot'' spread, ``ingot'' became the dominant spelling and is considered a correct word nowadays.

HeadFilt is applicable to text in different domains or to text written by different demographic groups because of its adaptability.
Theoretically, one could construct confusion sets automatically~\cite{chen2009word,chiu2014chinese,wang2018hybrid} for all characters.
In practice, it is difficult to construct ones that work well across different input methods (typed and written), different demographics (including new populations with specific slangs and vocabulary) and different domains.
Constructed confusion sets eventually get outdated overtime, so domain adaptation approaches are useful for a number of reasons.

Domain adaptation for filters requires estimating similarity between Chinese characters, which is algorithmically challenging because Chinese characters are not sequences of sub-character units; they have hierarchical structures.
Previous work estimated similarity between characters using handcrafted formulas based on edit distance~\cite{chen2013study,hong2019faspell}, Dice coefficients~\cite{liu2011visually}, or overlap in constituent structure~\cite{chang2013automatic}.
Edit distance was originally formulated for sequences while Dice coefficient was originally formulated for sets, so they might not capture the hierarchical nature of Chinese characters well.
Besides, since these measures are handcrafted, they likely require manual fine-tuning to work well.
In contrast, hierarchical embeddings can better characterize the similarity between Chinese characters and enable automated domain adaptation, reducing the manual effort to keep the filters up-to-date.

\begin{table}[t]
\small
\centering
\begin{tabular}{l@{\hspace{1em}}l@{\hspace{.8em}}l@{\hspace{.8em}}l@{\hspace{.8em}}l}
Correction & Acc. & Pre. & Rec. & F1 \\
\midrule
\multicolumn{5}{c}{2013 Bake-off} \\
\midrule
HeadFilt   & \cb 0.591 & 0.754 & --- & --- \\
-Ad        & 0.590 & 0.753 & --- & --- \\
-Ad -HE    & 0.577 & 0.755 & --- & --- \\
FixedFilt  & 0.575 & 0.737 & --- & --- \\
\midrule
\multicolumn{5}{c}{2014 Bake-off} \\
\midrule
HeadFilt   & \cb 0.698 & 0.811 & 0.517 & \cb 0.631 \\
-Ad        & 0.692 & 0.813 & 0.498 & 0.617 \\
-Ad -HE    & 0.691 & 0.813 & 0.496 & 0.616 \\
FixedFilt  & 0.684 & 0.816 & 0.475 & 0.601 \\
\midrule
\multicolumn{5}{c}{2015 Bake-off} \\
\midrule
HeadFilt   & \cb 0.746 & 0.885 & 0.565 & \cb 0.690 \\
-Ad        & 0.743 & 0.894 & 0.552 & 0.683 \\
-Ad -HE    & 0.742 & 0.899 & 0.546 & 0.679 \\
FixedFilt  & 0.729 & 0.890 & 0.523 & 0.658 \\
\end{tabular}
\caption{\label{tbl:ablation_corr}\textit{Traditional Chinese error correction ablation.
The means of 5 runs with different seeds are reported.
}}
\end{table}

\subsection{Better Recall with HeadFilt}
\begin{figure}[h]
\centering
\includegraphics[scale=0.3]{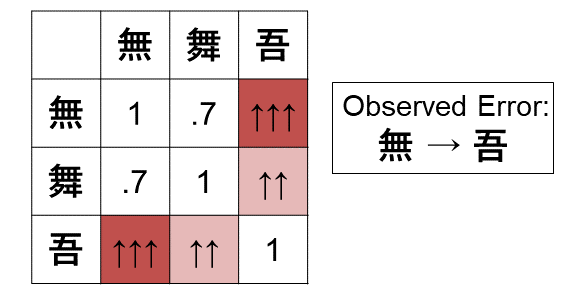}
\caption{\label{fig:discussion}\textit{Learning similarity without observation. Arrows indicate magnitude of change. Increasing the similarity score between \protect\begin{CJK*}{UTF8}{bsmi}吾\protect\end{CJK*} and \protect\begin{CJK*}{UTF8}{bsmi}無\protect\end{CJK*} indirectly increases the similarity score between \protect\begin{CJK*}{UTF8}{bsmi}吾\protect\end{CJK*} and \protect\begin{CJK*}{UTF8}{bsmi}舞\protect\end{CJK*}.}}
\end{figure}

The way hierarchical embeddings are optimized (Figure~\ref{fig:discussion}) could conceivably explain why HeadFilt achieves higher error correction recall than filtering using given confusion sets (FixedFilt) (Section~\ref{ssec:results} and~\ref{ssec:ablation}).
As \begin{CJK*}{UTF8}{bsmi}吾\end{CJK*} is a substitution error of \begin{CJK*}{UTF8}{bsmi}無\end{CJK*}, minimizing the contrastive loss increases their similarity and reduces the L2 distance between their embeddings.
Since \begin{CJK*}{UTF8}{bsmi}無\end{CJK*} and \begin{CJK*}{UTF8}{bsmi}舞\end{CJK*} have similar morphological structures, their hierarchical embeddings would be close in the embedding space.
Thus, increasing the similarity between \begin{CJK*}{UTF8}{bsmi}吾\end{CJK*} and \begin{CJK*}{UTF8}{bsmi}無\end{CJK*} would indirectly increase the similarity between \begin{CJK*}{UTF8}{bsmi}吾\end{CJK*} and \begin{CJK*}{UTF8}{bsmi}舞\end{CJK*}.
Even though \begin{CJK*}{UTF8}{bsmi}吾\end{CJK*} and \begin{CJK*}{UTF8}{bsmi}舞\end{CJK*} are not observed as substitution errors, by leveraging the morphological similarity, HeadFilt can still infer that there might be a substitution error.
Since many errors are not observed during training, using hierarchical embeddings widens the filter's coverage.

\subsection{Future Work}
Although HeadFilt models characters' similarity using their structures, which implicitly encode pronunciation~\cite{nguyen2018multimodal,nguyen2019hierarchical}, there may still be ambiguous cases.
Therefore, incorporating Pinyin or Zhuyin into the proposed HeadFilt could be a possible extension that further enhances similarity estimation and consequently improves spell checking.
This is an interesting possibility for future work, as the design and construction of such a study would require one to explicitly consider the contexts of the various different phonetic and tonal representations of a heteronym (a heteronym is a Chinese character that can take on multiple pronunciations), which requires extensive linguistic expertise and resources.
Such efforts are out of the scope of this current endeavor.

Although this work focuses on spell check for Chinese, it can be extended to languages with similar logographic and phonological structures such as Japanese, Korean, or Vietnamese~\cite{ngo2014minimal,ngo2019phonology}.
Besides spell check, other tasks such as automatic speech recognition may also achieve higher recall by using HeadFilt.
There are many homophones in Chinese languages and predicting the correct characters from the audio signal is at times challenging.
HeadFilt can narrow the list of candidate characters with similar pronunciation to lower the perplexity for speech recognition.

\section{Conclusion}
We presented a filtering model using hierarchical character embeddings (HeadFilt) for Chinese spell check.
Our approach achieved SOTA error correction results on two spell check datasets.
The adaptability of HeadFilt makes it possible to tailor proposed model to text from different domains, text written using different input methods, and potentially text written by different individuals.

\ifCLASSOPTIONcaptionsoff
  \newpage
\fi

\bibliographystyle{IEEEtran}
\bibliography{spell,nlp}

\end{document}